\documentclass[runningheads]{llncs}

\usepackage{accv}

\usepackage{accvabbrv}

\usepackage{graphicx}
\usepackage{booktabs}

\usepackage[accsupp]{axessibility}  

\usepackage{amsmath}
\usepackage{amssymb} 
\usepackage{dsfont}

\usepackage{graphicx}
\usepackage{amsfonts}
\usepackage{booktabs}
\usepackage{multicol}
\usepackage{multirow}
\usepackage{xcolor} 
\usepackage{soul}
\usepackage{subcaption}
\usepackage{xspace}
\usepackage{lipsum}
\usepackage{amssymb}
\usepackage{pifont}
\newcommand{\cmark}{\ding{51}}%
\newcommand{\xmark}{\ding{55}}%
\newcommand{\model}{Gen-XAI\xspace}
\newcommand{\dataset}{FG-CXR\xspace}

\usepackage{enumitem}
\setlist{nosep, leftmargin=14pt}
\usepackage{soul}

\usepackage[pagebackref,breaklinks,colorlinks,citecolor=accvblue]{hyperref}
\usepackage[capitalize]{cleveref}
\Crefname{section}{Section}{Sections}
\Crefname{table}{Table}{Tables}
\Crefname{figure}{Figure}{Figures}
\usepackage{orcidlink}

\begin{document}

\title{FG-CXR: A Radiologist-Aligned Gaze Dataset for Enhancing Interpretability in Chest X-Ray Report Generation} 

\titlerunning{FG-CXR}
\authorrunning{Pham et al.}
\author{Trong Thang Pham*\inst{1} \and Ngoc-Vuong Ho\inst{1} \and Nhat-Tan Bui\inst{1} \and Thinh Phan\inst{1},\\
Patel Brijesh\inst{3} \and Donald Adjeroh\inst{3} \and Gianfranco Doretto\inst{3} \and Anh Nguyen\inst{4} \and \\ Carol C. Wu\inst{5} \and Hien Nguyen\inst{2} 
\and Ngan Le\inst{1}}

\institute{$^{1}$University of Arkansas, Fayetteville, AR, USA \\ 
$^{2}$University of Houston, Houston, Texas, USA \\ 
$^{3}$West Virginia University, Morgantown, West Virginia, USA \\ 
$^{4}$University of Liverpool, UK \\ 
$^{5}$MD Anderson Cancer Center, Houston, Texas, USA \\ 
*Corresponding author: \email{tp030@uark.edu}}

\maketitle

\begin{abstract}
Developing an interpretable system for generating reports in chest X-ray (CXR) analysis is becoming increasingly crucial in Computer-aided Diagnosis (CAD) systems, enabling radiologists to comprehend the decisions made by these systems. Despite the growth of diverse datasets and methods focusing on report generation, there remains a notable gap in how closely these models's generated reports align with the interpretations of real radiologists. In this study, we tackle this challenge by initially introducing \emph{Fine-Grained CXR} (\dataset) dataset, which provides fine-grained paired information between the captions generated by radiologists and the corresponding gaze attention heatmaps for each anatomy. Unlike existing datasets that include a raw sequence of gaze alongside a report, with significant misalignment between gaze location and report content, our FG-CXR dataset offers a more grained alignment between gaze attention and diagnosis transcript.
Furthermore, our analysis reveals that simply applying black-box image captioning methods to generate reports cannot adequately explain which information in CXR is utilized and how long needs to attend to accurately generate reports. Consequently, we propose a novel \emph{explainable radiologist’s attention generator} network (\model) that mimics the diagnosis process of radiologists, explicitly constraining its output to closely align with both radiologist's gaze attention and transcript. Finally, we perform extensive experiments to illustrate the effectiveness of our method. Our datasets and checkpoint is available at \url{https://github.com/UARK-AICV/FG-CXR}.
\end{abstract}

\section{Introduction}
\label{sec:intro}
Chest X-rays (CXRs) are commonly used for both screening and diagnostic purposes, resulting in a substantial daily workload. Additionally, the current shortage of trained radiologists in many healthcare systems highlights the need for automated radiology report generation to help reduce radiologists' workloads~\cite{wuchest}. The success of Deep Learning~\cite{rudin2019stop,le2024infty,nguyen2024tackling,nguyen2024contrastive,nguyen2024occluded,le2023music,tran2020itask,pham2024style,vo2023dna,coffman2024cattleface} has led people to pursue its application in the medical domain~\cite{nguyen2023embryosformer,bigolin2022reflacx}. However, most existing methods lack explainability, which is a major reason for their limited adoption. In the safety-critical medical field, a highly accurate but opaque report generation system may not be adopted if the reasoning behind the generated report is not transparent and explainable~\cite{geis2019ethics,guidotti2018survey,miller2019explanation}. Therefore, creating and using an interpretable system should be prefer to black-box system~\cite{rudin2019stop}.

In the examination process, radiologists carefully examine every anatomy of CXRs and report their findings. Inspired by this process, we hypothesize that understanding pixel importance and gaze patterns can improve AI model explainability and accuracy in CXR diagnosis. However, the use of radiologist gaze-derived heatmaps in generating descriptive reports during CXR diagnosis remains underexplored. Recently, Tanida et al.~\cite{tanida2023rgrg} address this challenge by introducing an interpretable system that uses bounding boxes, which lack detail. In contrast, Pham et al.~\cite{pham2024ai} propose a diagnosis system directly supervised by gaze attention. However, this system is limited as it can only predict whether an anatomical region is abnormal, requiring users to identify the specific findings themselves, which can be impractical.

\begin{figure}[!t]
    \centering
    \includegraphics[width=\linewidth]{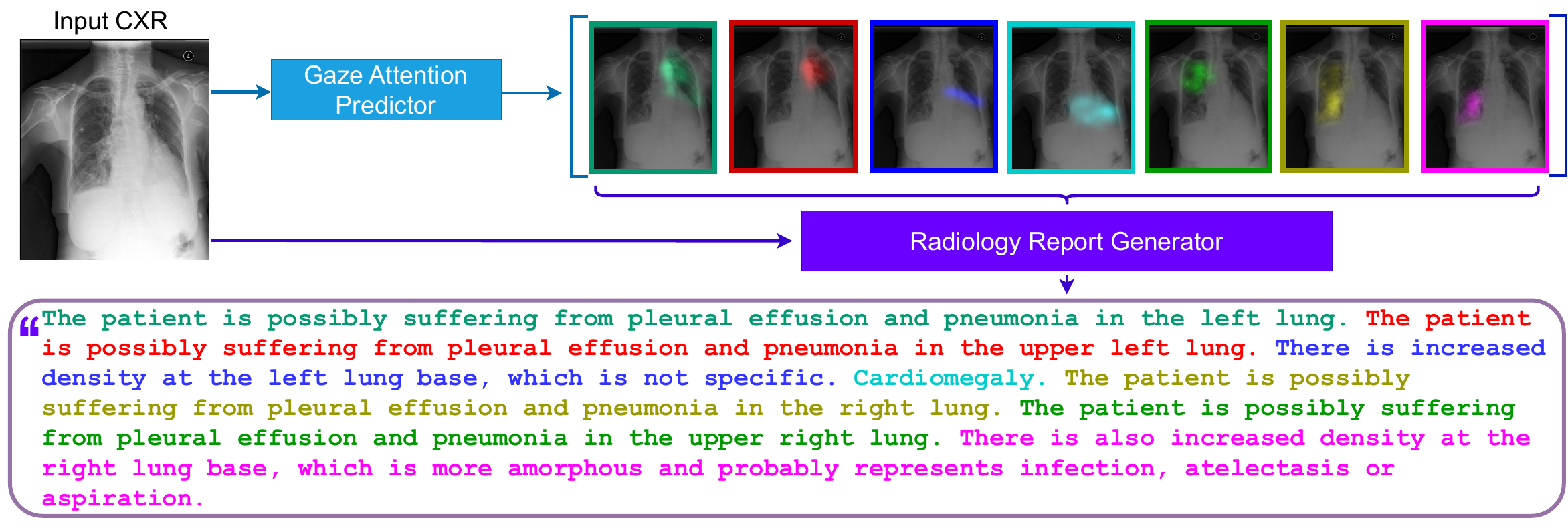}
    \caption{An overview of our interpretable \model framework, generating a diagnosis report and its corresponding visual attention for each diagnosis in the report. }
    \label{fig:overview}
\end{figure}

To address the aforementioned weaknesses, we introduce \model pipeline, shown in \cref{fig:overview}. \model mimics how radiologists perceive images by decoding radiologist's gaze attention with the Gaze Attention Predictor and then explaining its observations through the Report Generator. The Gaze Attention Predictor focuses on learning the regions of interest based on radiologists' gaze attentions, ensuring that the system captures the critical areas that a radiologist would typically examine. The Report Generator then uses this information to produce an accurate radiology report, which is visually grounded with the anatomical gaze attention, enhancing the transparency and explainability of the diagnostic process.

Existing gaze datasets~\cite{bigolin2022reflacx,karargyris2020eye} provide raw gaze sequences along with reports for each patient. However, radiologists typically observe before diagnosing, leading to a misalignment between the gaze location and the report at the same timestamp, as illustrated in \cref{fig:Data_comparison}. Therefore, a cleaner dataset is needed to evaluate this pipeline effectively. To address this, we curate a new dataset that provides gaze sequences aligned with anatomical attention heatmaps. By aligning gaze sequences with attention heatmaps, we ensure that the generated reports are not only accurate but also provide insights into the reasoning process behind each diagnosis.
Our main contributions are summarized as follows:
\begin{itemize} 
\item We introduce \dataset, a curated dataset that provides anatomical segmentation, gaze attention heatmaps annotated by radiologists, and radiology reports that are aligned with the gaze attention heatmaps.  
\item We propose a novel interpretable baseline \model to efficiently generate radiology reports with meaningful attention heatmap. 
\end{itemize}





\section{Related works}
\subsection{Interpretable Deep Learning.} 
In high-stakes medical settings, understanding the decision-making process is crucial~\cite{rudin2022interpretable}. A direction to enhance interpretability is to design an architecture that can learn concepts~\cite{kim2018interpretability,pham2024ai}. \textit{In our paper, we follow the interpretable approach~\cite{rudin2022interpretable} by learning radiologists' intentions (gaze attention) across anatomical parts. However, unlike previous works~\cite{selvaraju2017grad,pham2024ai,nauta2023pip} focusing on classification, we address the less explored task: the model must explain observations via report generation based on inferred intentions.
}

\subsection{Interpretable-oriented Datasets.}
Creating datasets with annotated abnormality localization traditionally involves manual curation, but this is labor-intensive and often yields limited coverage, typically 1-2 labels ~\cite{filice2020crowdsourcing,shih2019augmenting}. Recent efforts address this by providing datasets with anatomy labels in reports. For instance, the Chest ImaGenome dataset ~\cite{wuchest} is a dataset containing localized annotations (bounding box), with corresponding reports for the associated CXR images. However, existing datasets lack the granularity needed, e.g. gaze, to develop models that mimic real radiologist diagnoses. \textit{In contrast, our dataset enhances detail by mapping reports to 7 anatomical locations using radiologist attention heatmaps, providing deeper insights into the diagnostic process. } 

\begin{table}[t!]
\centering
\caption{
Overview of CXR datasets. A coarse report describes the whole image. A fine-grained report is a report associated with individual anatomies.}
\label{tab:related_datasets}
\resizebox{\textwidth}{!}{%
\begin{tabular}{@{}c|l|l|l|r|c@{}}
\toprule
\multirow{2}{*}{\textbf{Gaze}} & \multicolumn{1}{c|}{\multirow{2}{*}{\textbf{Datasets}}} & \multicolumn{2}{c|}{\textbf{Annotation}}& \multirow{2}{*}{\textbf{\#Samples}} & \multirow{2}{*}{\textbf{NLP Reports}} \\ \cline{3-4}
& & \textbf{Information} & \textbf{Method} & & \\ \midrule
\multirow{12}{*}{\xmark} 
& SIIM-ACR Pneumothorax Segmentation \cite{filice2020crowdsourcing} & Segmentation & Manual + augmented  &  12,047  & No \\
& RSNA Pneumonia Detection Challenge   \cite{shih2019augmenting} & Bounding Boxes & Manual &  30,000 & No  \\
& NIH CXR dataset \cite{wang2017chestx} & Entire CXR & Automated  &   112,120  & No \\
& PLCO \cite{team2000prostate} & Entire CXR & Automated  &  236,000  & No \\
& Stanford CheXpert \cite{irvin2019chexpert} & Entire CXR & Automated  &  224,316  & No \\
& Montgomery County Chest X-ray   \cite{jaeger2014two} & Segmentation & Manual & 138  & No \\
& Shenzen Hospital Chest X-ray   \cite{jaeger2014two} & Segmentation & Manual    & 662  & No \\  
& Indiana University Chest X-ray Collection \cite{demner2016preparing} & Entire CXR & Automated & 3,813 & Coarse \\
& MIMIC-CXR \cite{johnson2019mimic} & Entire CXR & Automated  &  377,110  & Coarse \\
& Dutta \cite{datta2020dataset} & Entire CXR & Manual  &  2,000  & Coarse \\
& PadChest \cite{bustos2020padchest} & Entire CXR & Manual + automated   & 160,868  & Coarse \\
& Chest ImaGenome \cite{wuchest} & Bounding Boxes & Automated  &  242,072 & Fine-Grained  \\ \midrule
\multirow{3}{*}{\cmark} & 
REFLACX \cite{bigolin2022reflacx} & Gaze & Automated  & 3,000 & Coarse \\ 
& EGD \cite{karargyris2020eye} & Segmentation + Gaze & Automated &  1,000 & Coarse \\ 
\cline{2-6}
& \multirow{3}{*}{\textbf{Our \dataset}} & Atanomies Localization & {Semi-automated} &
\multirow{3}{*}{2,951} & \multirow{3}{*}{Fine-Grained} \\  
&&  Gaze Attention Heatmap   & Automated &  & \\  
&&  Gaze Sequence  & Automated &   & \\ 
\bottomrule
\end{tabular}%
}
\end{table}

\subsection{Radiology Report Generation.}
Early approaches~\cite{jing2019show,li2018hybrid} in report generation leveraged CNN-RNN architectures or transformer inspired by general image captioning. However, medical report generation differs from image captioning~\cite{tanida2023rgrg} due to varying lengths, complexities, and biases in normal samples. To address these challenges, some models align visual features with disease tags~\cite{you2021aligntransformer}, while others incorporate medical knowledge graphs~\cite{liu2021exploring}. Notably, RGRG~\cite{tanida2023rgrg} tackles interpretability by outlining abnormal regions and generating captions about them, but it lacks precision in specifying abnormality areas within bounding boxes. \textit{In contrast, our method simulates radiologists' focus on important regions and generates insights based on them.}

\begin{figure}[!thb]
    \centering
    \includegraphics[width=0.9\linewidth]{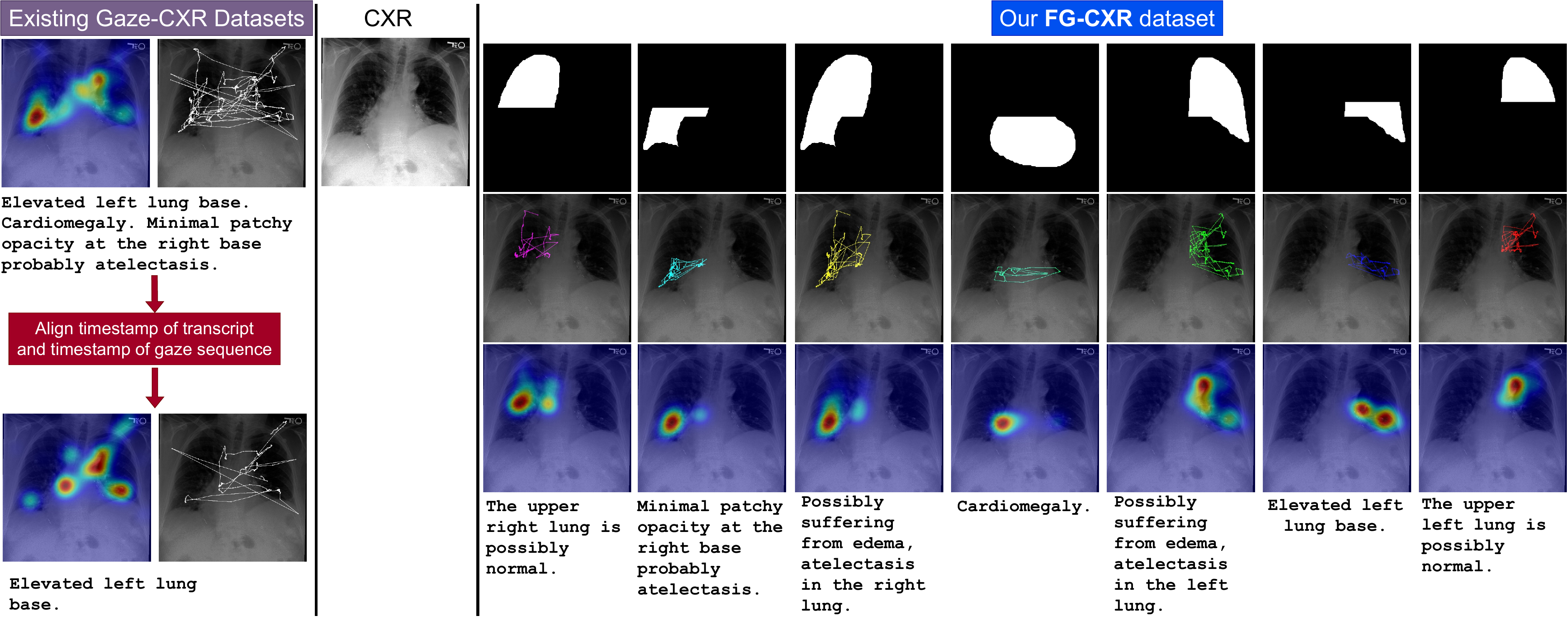}
    \caption{Annotation comparison between prior gaze-CXR datasets (left) which face challenges in aligning gaze location with textual description and our \dataset (right), given a CXR (middle). }
    \label{fig:Data_comparison}
\end{figure}

Our \dataset dataset closely simulates radiologists' real-life diagnostic process by providing detailed annotations, including anatomical localization, gaze attention, and corresponding medical reports. A comparison between our \dataset with the existing CXR datasets is given in Table \ref{tab:related_datasets}, while a visualization of the comparison of gaze-based annotations is shown in Fig. \ref{fig:Data_comparison}. 

\section{Dataset: Fine-Grained CXR (\dataset)}
\label{sec:data}


\subsection{Anatomy Localization.} According to radiologists, their focus can be divided into seven key areas of CXR: heart, left, right, upper left, upper right, lower left, and lower right lungs. Therefore, we create anatomical masks for seven regions. Leveraging CXRs and gaze sequences from EGD~\cite{karargyris2020eye} and REFLACX~\cite{bigolin2022reflacx}, we apply techniques from \cite{pham2024ai} to generate detailed masks for the heart and lungs, segmented into upper and lower regions. Finally, images with extreme brightness are filtered out. 
\begin{table}[!tb]
\centering
\caption{\textbf{Keywords for Anatomical Regions.} The keywords are chosen by listing all sentences from the radiology reports and picked based on their meaning.}
\label{tab:keywords}
\resizebox{\textwidth}{!}{%
\begin{tabular}{l|l}
\hline
\textbf{Anatomical Region} & \multicolumn{1}{c}{\textbf{Keywords}} \\
\midrule
\texttt{Heart} & \texttt{cardiomegaly}, \texttt{enlarged} and \texttt{chest}, \texttt{heart}, \texttt{cardiac}, \texttt{mediastinum} \\
\texttt{Left Lung} & \texttt{left} \\
\texttt{Right Lung} & \texttt{right} \\
\texttt{Upper Left Lung} & \texttt{upper} and \texttt{left}, \texttt{apex} and \texttt{left}, \texttt{mid} and \texttt{left}, \texttt{apical} and \texttt{left}, \\
 & \texttt{top} and \texttt{left} \\
\texttt{Upper Right Lung} & \texttt{upper} and \texttt{right}, \texttt{apex} and \texttt{right}, \texttt{mid} and \texttt{right}, \texttt{apical} and \texttt{right},\\
 &  \texttt{top} and \texttt{right} \\
\texttt{Lower Left Lung} & \texttt{lower} and \texttt{left}, \texttt{base} and \texttt{left}, \texttt{bottom} and \texttt{left} \\
\texttt{Lower Right Lung} & \texttt{lower} and \texttt{right}, \texttt{base} and \texttt{right}, \texttt{bottom} and \texttt{right} \\
\hline
\end{tabular}}
\end{table}

\subsection{Anatomical-aware Gaze Attention.} 
\label{sec:gaze_attention_filtering}
Given the gaze coordinates $G = \{g_1, g_2,$ $ \dots, g_{|G|}\} \in \mathbb{N}^{|G| \times 2}$ of a CXR, our filtering process as the follows: For a report $T = \{s_1, s_2, ... s_{|T|}\} $, we identify all $s_i$ that include keywords pertinent to the area of interest, select the latest end time and remove all gaze points after that timestamp. The list of keywords is described in \cref{tab:keywords}. If transcript lacks keywords indicating anatomy, we use the entire gaze sequence. Finally, we filter out any gaze points that fall outside the segmentation masks corresponding to the anatomical areas of interest. The final gaze sequence of a $i^{th}$ sample is represented in two forms:  gaze sequence in a temporal order $G^{i} \in \mathbb{N}^{|G^{i}| \times 2} \subseteq G $, and gaze attention heatmap $A \in \mathbb{R}^{H \times W}$, which is created by creating gaze frequency map and applying Gaussian blurring as in \cite{karargyris2020eye}.

\subsection{Anatomical-aware Report}

For every anatomical region, we associate it with a brief report. For instance, we link ``the heart is normal'' with the heart area. However, a report from REFLACX~\cite{bigolin2022reflacx} or EGD~\cite{karargyris2020eye} might only include certain anatomical regions. To address this, we use a template to generate reports for any missing anatomy. If a diagnosis for any region is absent after keyword filtering (\cref{sec:gaze_attention_filtering}), we create a default sentence based on MIMIC-CXR annotations: \texttt{``the \{area\} is possibly normal''} for no findings, or \texttt{``the patient is possibly suffering from \{findings\} in the \{area\}''} for specific findings. For example, if a patient's current report lacks information for the left lung area, we refer to MIMIC-CXR and find that the label for this patient is ``no finding''. We then generate the sentence ``the left lung is possibly normal'' for this patient's left lung.

\subsection{Dataset Statistics}
\begin{figure}[!tb]
  \centering
    \begin{subfigure}[b]{0.45\textwidth}
        \includegraphics[width=\textwidth]{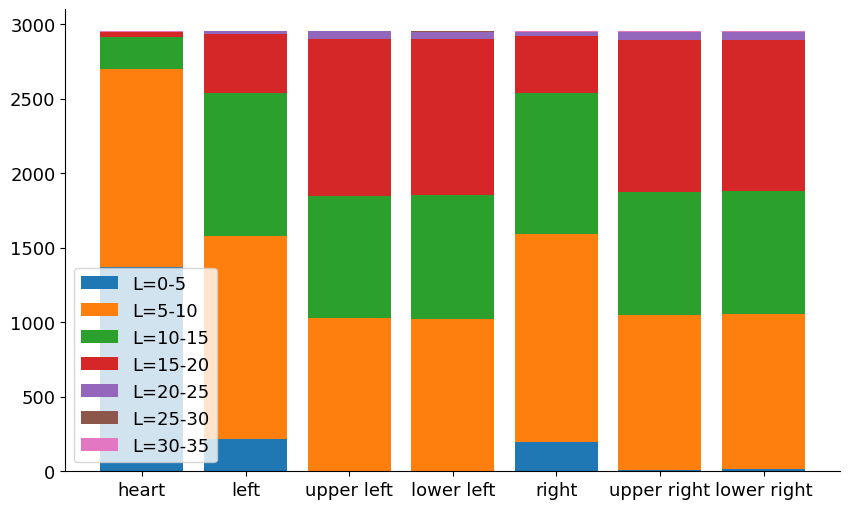}
        \caption{\textbf{Regional report lengths.} }
        \label{fig:data_stat_sub1}
    \end{subfigure}
    \hfill 
    \begin{subfigure}[b]{0.45\textwidth}
        \includegraphics[width=\textwidth]{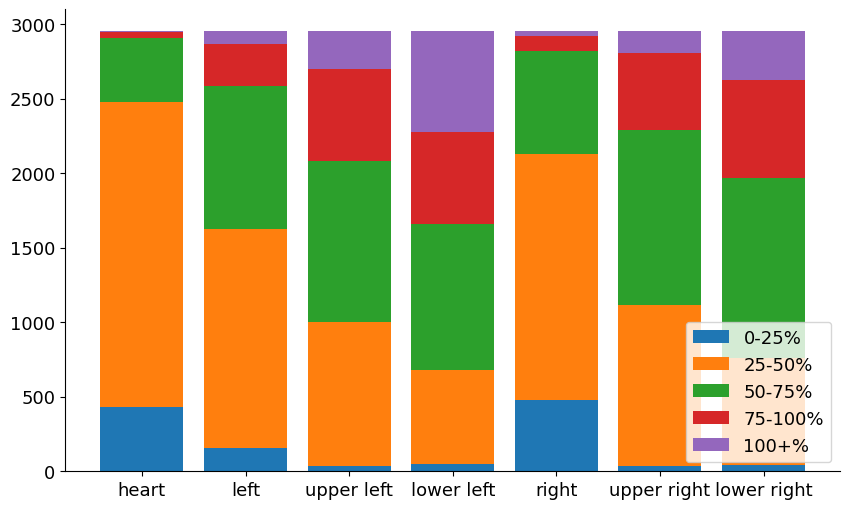}
        \caption{\textbf{Heatmap to mask area ratios.} }
        \label{fig:data_stat_sub2}
    \end{subfigure}
    \\
    \begin{subfigure}[b]{0.45\textwidth}
        \includegraphics[width=\textwidth]{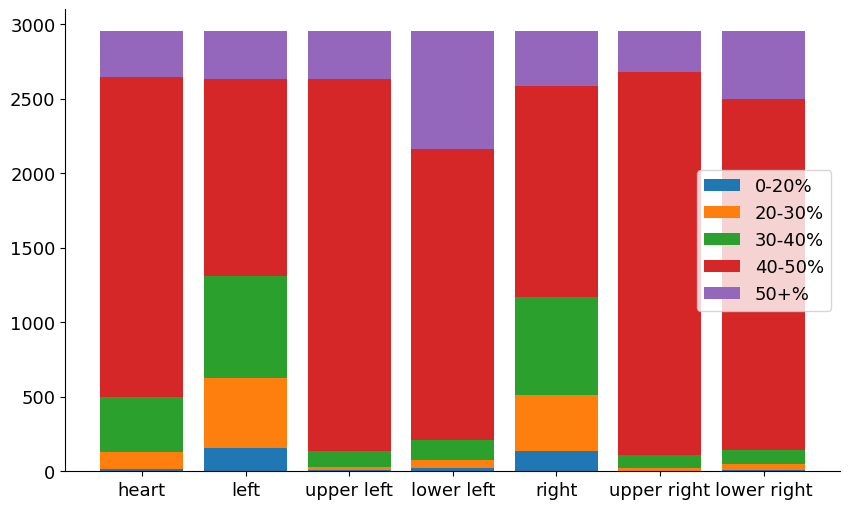}
            \caption{\textbf{Heatmap to bounding box area ratios.} }
        \label{fig:data_stat_sub3}
    \end{subfigure}
    \hfill
    \begin{subfigure}[b]{0.45\textwidth}
        \includegraphics[width=\textwidth]{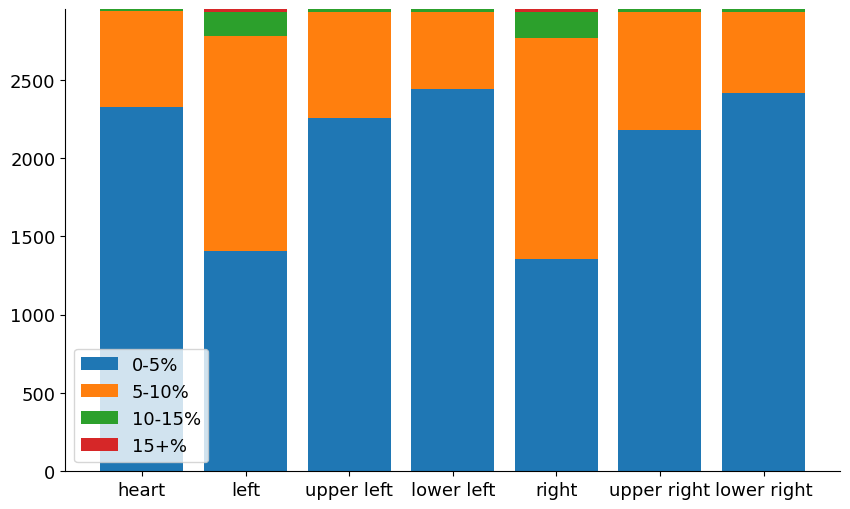}
        \caption{\textbf{Heatmap to full image size ratios per region.} }
        \label{fig:data_stat_sub4}
    \end{subfigure}
  \caption{Dataset distributions for \dataset.}
  \label{fig:data_stat}
\end{figure}

Our dataset contains 2,951 CXRs in total, with 20,657 pairs of \{attention heatmap, report\}. \cref{fig:data_stat} provides deeper insights into our dataset. 

\noindent
\textbf{Reports are mostly concise.}
In \cref{fig:data_stat_sub1}, we observe that the majority of reports are concise, particularly those describing the heart, with most reports comprising fewer than 10 words. Reports on other anatomical regions tend to be longer, though a significant number still fall within the 5 to 10 words range. This indicates a prevalent practice among radiologists of using succinct sentences.

\noindent
\textbf{Radiologist's attention versus segmentation mask.}
\cref{fig:data_stat_sub2} reveals that most gaze heatmaps typically cover a smaller area than the full anatomical segmentation mask. Notably, many heatmaps in the lower left and right lungs encompass a larger area (resulting in a ratio greater than 1). This can be attributed to the presence of dense gaze sequences that extensively cover a particular region, and this is further amplified with the Gaussian filtering process. Such a phenomenon is particularly evident in the lower right and lower left regions, likely due to radiologists' meticulous examination of the diaphragm, which extends beyond the lower masks. Furthermore, the base of the left lung mask is typically smaller than that on the right side, attributed to occlusion by the heart. 

\noindent
\textbf{Radiologist's attention versus bounding box.}
In \cref{fig:data_stat_sub3}, most heatmaps are confined to a portion of their bounding box, which is computed by taking the top left and bottom right corners of the non-zero heatmap values. A notable observation is that many gaze attention heatmaps utilize less than 20\% of the bounding box area, especially in the left and right lungs.

\noindent
\textbf{Radiologist's attention versus the whole image.}
In \cref{fig:data_stat_sub4}, most heatmaps use little information from the whole image. This is true even for the left and right lungs, where one might expect a higher coverage area; however, most heatmaps occupy only about 10\% of the image's total area.

\begin{table}[!tb]
\centering
\caption{\textbf{Dataset splits for training, validation, and testing sets.} From the data in \cref{sec:data}, we create these splits for training, validating, and testing our method in \cref{sec:exp}.}
\label{tab:dataset-distribution}
\begin{tabular}{l|c|c}
\toprule
\textbf{Set} & \textbf{Number of Samples} & \textbf{Percentage} \\
\midrule
{Training Set} & 2,074 & 70\% \\
{Validation Set} & 295 & 10\% \\
{Testing Set} & 582 & 20\% \\
\hline
\textbf{Total} & 2,951 & \textbf{100\%} \\
\bottomrule
\end{tabular}
\end{table}

For benchmarking in \cref{sec:exp}, we randomly split our dataset into 70\% for training, 10\% for validation, and 20\% for testing. The number of samples is shown in \cref{tab:dataset-distribution}.

\subsection{How will our \dataset benefit the community?} 
We anticipate that the release of our \dataset will drive advancements in these promising areas:
\begin{itemize}
    \item Gaze-Interpretable Report Generation: While report generation is a growing research topic, enhancing and evaluating report generation with explainability using radiologist gaze data remains relatively unexplored.
    \item General Medical Tasks: The \dataset dataset, enriched with segmentation masks for critical anatomical areas, serves as a valuable resource for developing and benchmarking Anatomical Segmentation algorithms \cite{lei2020self,ullah2023deep}. The reports of \dataset dataset, validated by experts, is also richer and more informative than the original REFLACX \cite{bigolin2022reflacx} and EGD \cite{karargyris2020eye} datasets, making it a potential benchmark for Radiology Report Generation \cite{zhang2020radiology,liu2021exploring}.
\end{itemize}

\section{Methodology}
\label{sec:arch}

\begin{figure*}[!t]
    \centering
    \includegraphics[width=\linewidth]{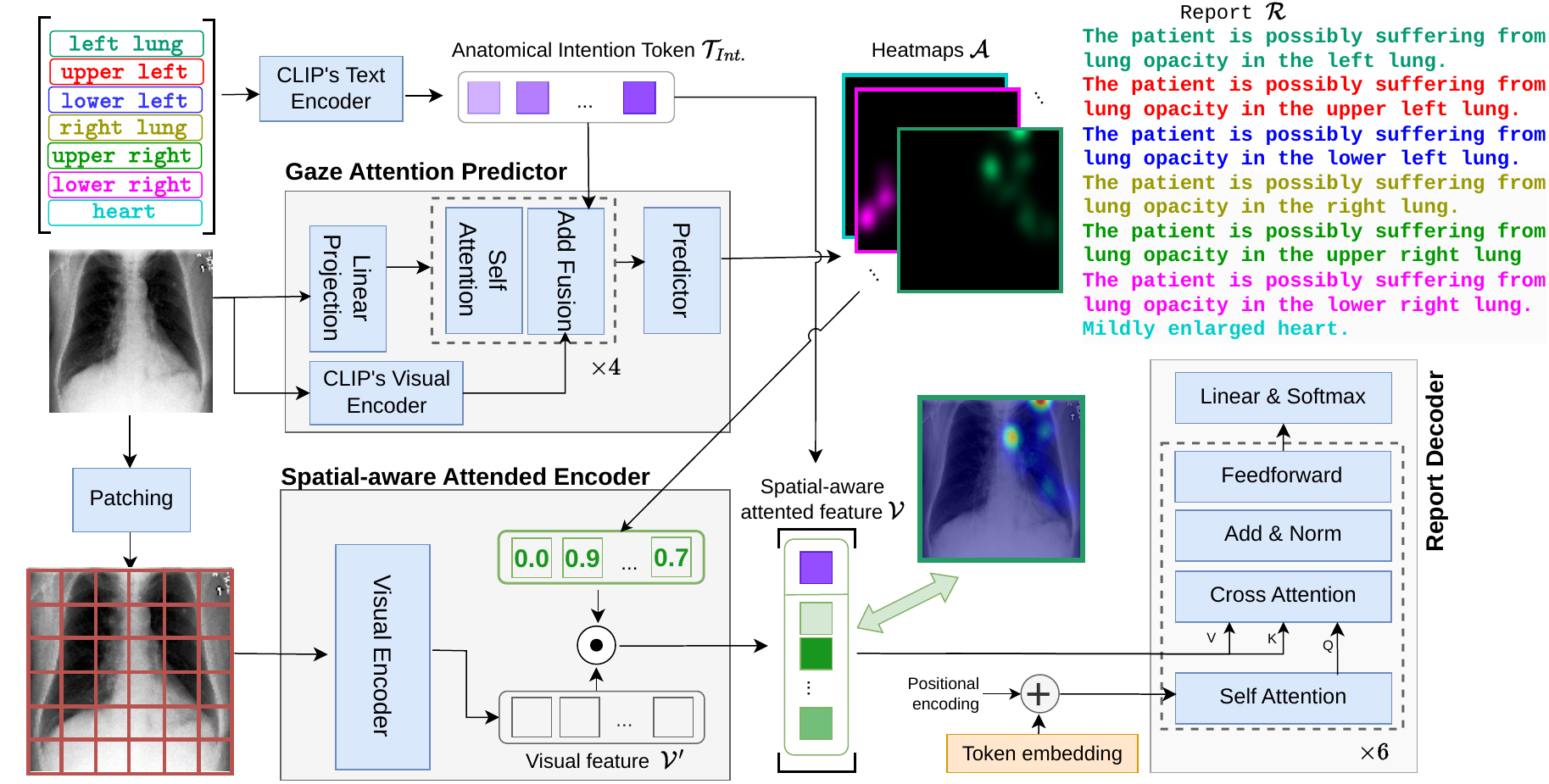}
    \caption{The detailed architecture of our framework consisting of three key modules: Gaze Attention Predictor, Spatial-Aware Attended Encoder, and Report Decoder.}
    \label{fig:detail}
\end{figure*}




In this section, we introduce a novel framework for Gaze-interpretable Report Generation.
Given a CXR image $\mathcal{I}$, our goal is to produce gaze-based heatmaps $\mathcal{A}$ and generate a report $\mathcal{R}$ of the attended region $\mathcal{A}$. To ensure interpretability, the gaze attention $\mathcal{A}$ has to be closely aligned with expert observation and the report $\mathcal{R}$ has to be consistent with what gaze attention $\mathcal{A}$. Our \model architecture, detailed in \cref{fig:detail}, comprises three key modules: (i) Gaze Attention Predictor (GAP) to predict seven gaze-based heatmaps $\mathcal{A}$; (ii) Spatial-aware Attended Encoder (SAP) to generate attended features $\mathcal{V}$; and (iii) Report Decoder to generate diagnosis reports for seven anatomies.


\subsection{Gaze Attention Predictor (GAP)} Given a CXR image $\mathcal{I}$ and Anatomical Intention Token $\mathcal{T}_{Int.}$, this module predicts the radiologist-like attention heatmap $\mathcal{A} \in [0,1]^{7 \times H/16 \times W/16 \times 1}$. 
Inspired by \cite{pham2024ai}, we adopt the idea of training a deep adapter on a pretrained CLIP \cite{radford2019language} to predict heatmaps. Initially, we extract four intermediate features from the middle layers (i.e. $\{0,3,6,9\}$ according to \cite{pham2024ai}) of the CLIP's visual encoder from the $\mathcal{I}$. Subsequently, $\mathcal{I}$ is split into $H/16 * W/16$ patches with size $16 \times 16$ and projected into an embedding space $\mathcal{I}_e$ to prepare for the fusion stage via a Linear Projection layer. In the fusion stage, we fuse $\mathcal{I}_e$ with the extracted CLIP's visual features $\mathcal{V}$ and $\mathcal{T}_{Int.}$. Finally, a Predictor module, consisting of an MLP followed by a sigmoid activation, predicts seven gaze heatmaps for the seven parts of the lung.
To create $\mathcal{T}_{Int.}$, we use CLIP's text encoder to extract seven textual features from:  \texttt{``heart''}, \texttt{``left''}, \texttt{``right''}, \texttt{``upper left''}, \texttt{``upper right''}, \texttt{``lower left''}, and \texttt{``lower right''}; and we stack them into a tensor $\mathcal{T}_{Int.}$ of size $7 \times D$. This allows simultaneous prediction of all seven parts, rather than individually.

\subsection{Spatial-aware Attended Encoder (SAP)} This module produces an attended feature $\mathcal{V}$ that contains information from its patch and neighbors, crucial for focusing on relevant areas while retaining essential spatial context. 
Given CXR's grayscale nature, distinguishing areas like the lung from the background requires understanding their spatial relation to adjacent patches. This spatial information only exists in the encoded feature~\cite{wu2021cvt,le2024infty}.  Unlike previous works~\cite{pham2024ai,kashyap2020looking} that apply attention to pixel level that may remove vital details, our approach applies attention to the latent visual features $\mathcal{V}'$. The effectiveness is empirically proven and included in \cref{sec:ablation}. Specifically, we first extract the patches' visual feature $\mathcal{V}' \in \mathbb{R}^{H/16 \times W/16 \times D}$ by using a Visual Encoder (i.e. CvT~\cite{wu2021cvt}). Then, we create the spatial-aware attended feature by performing element-wise multiplication between $\mathcal{A}$ and $\mathcal{'}$ to create the reweighted feature $\mathcal{V}$. To further guide the model, we also concatenate the token of the current area of interest into the feature, for example concatenating the intention token of looking at the heart to the feature that is masked by the heart heatmap. Mathematically, we compute $\mathcal{V} \in \mathbb{R}^{7 \times (H/16 * W/16 +1) \times D}$ with $\mathcal{V}(i) = [\mathcal{V'} \odot \mathcal{A}(i), \mathcal{T}_{Int.}(i)], \forall i \in [0,6]$, where $i$ indicates $i^{th}$ region, $[\cdot]$ is concatenation and $\odot$ is the Hadamard product. 

\subsection{Report Decoder} We utilize GPT2~\cite{radford2019language}, an auto-regressive network for text generation, as our textual report decoder architecture. The token embedding is the embedding of previous tokens, for example, \texttt{``[BOS],the,patient,is, possibly,suffer,from''} to predict the next word \texttt{``lung''}, where \texttt{[BOS]} is the beginning of sentence token. For every area $i$, we use $\mathcal{V}(i)$ as key (\texttt{K}) and value(\texttt{V}), and the output feature from self attention of token embedding as query (\texttt{Q})of the cross-attention module. After predicting all sentences, we concatenate them to create the final report.


\subsection{Learning Objective} 
We train our model with the training loss $\mathcal{L}= (1+\lambda_c)\mathcal{L}_c + (1+\lambda_h)\mathcal{L}_h$, where \(\lambda_c,\mathcal{L}_c, \lambda_h,\mathcal{L}_h  \) are the report penalty, cross-entropy loss for the generated report, heatmap penalty, and \(L_2\) loss for predicted heatmap, respectively.
To enhance the model's focus on predicting correct anatomy, we introduce two dynamic coefficients as penalties: gaze attention prediction ($\lambda_h$) and report generation ($\lambda_c$). During the heatmap prediction, we use Intersection over Union (IoU) with a threshold of 0.5 to identify instances where the model inaccurately focuses. Each incorrect prediction increases $\lambda_h$ by 1. $\mathcal{A}_{gt}$ is the ground truth gaze attention map. 
\begin{equation}
\lambda_h = \sum_{i} \mathds{1}_{0.5}(\operatorname{IoU}(\mathcal{A}(i),\mathcal{A}_{gt}(i)))
\text{, where } \mathds{1}_{0.5} = \begin{cases} 
1 & \text{if } x \geq 0.5 \\
0 & \text{otherwise}
\end{cases}
\end{equation}

At the report generation, we want the model the model explicitly predict directions while minimizing incorrect directional words. Thus, if the model fails to predict anatomical keywords or mentions the wrong direction, $\lambda_c$ increases by 1. For every CXR, both $\lambda_h$ and $\lambda_c$ are initialized to 0, indicating that they are not accumulated across all samples in an epoch. The ablation study on the effect of penalty terms is included in \cref{sec:ablation}.

\section{Experiments}
\label{sec:exp}

\begin{table}[!t]
\centering
\setlength{\tabcolsep}{5pt}
\renewcommand{\arraystretch}{0.9}
\caption{Performance comparison between our method and other SOTA methods on natural language generation (NLG) metrics for the full report generation task.}
\resizebox{\linewidth}{!}{
\begin{tabular}{l|ccccccc|cc}
\hline
        Methods & B1$\uparrow$ & B2$\uparrow$ & B3$\uparrow$ & B4$\uparrow$ & M$\uparrow$ & R$\uparrow$ & C$\uparrow$ & Div@2$\uparrow$ & R@4$\downarrow$\\ \hline
        R2Gen \cite{chen2020generating} &  0.690 & 0.576 & 0.502 & 0.459 & 0.326 & 0.607 & 3.302 & 0.357 & 0.459 \\ 
        R2GenCMN \cite{chen-etal-2021-cross-modal} & 0.688 & 0.572 & 0.513 & 0.472 & 0.326 & 0.612 & 3.355 & 0.437 & 0.174 \\
        CvT2DistilGPT2 \cite{nicolson_improving_2023} & 0.708 & 0.647 & 0.585 & 0.552 & 0.352 & 0.618 & 2.936 & 0.486 & 0.149 \\
        RGRG \cite{tanida2023interactive} &  0.715 & 0.598 & 0.583 & 0.550 & 0.351 & 0.532 & 3.300 & 0.624 & 0.075 \\ 
        $\mathcal{M}^2$ Transformer \cite{cornia2020m2} & 0.694 & 0.613 & 0.533 & 0.476 & 0.333 & 0.623 & 3.097 & 0.672 & 0.065 \\  \hline
        Ours & \textbf{0.729} & \textbf{0.658} & \textbf{0.606} & \textbf{0.561} & \textbf{0.386} & \textbf{0.692} & \textbf{4.026} & \textbf{0.854} & \textbf{0.055} \\ \hline
\end{tabular}}
\label{tab:nlg}
\end{table}

\begin{table}[!t]
\centering
\setlength{\tabcolsep}{5pt}
\renewcommand{\arraystretch}{0.9}
\caption{Performance comparison between our method and other SOTA methods on clinical efficacy (CE) metrics.}
\resizebox{\linewidth}{!}{
\begin{tabular}{l|ccc|ccc|ccc}
\hline
        Methods & P$_{\text{mic}}$ $\uparrow$ & R$_{\text{mic}}$$\uparrow$ & F1$_{\text{mic}}$$\uparrow$ & P$_{\text{mac}}$$\uparrow$ & R$_{\text{mac}}$$\uparrow$ & F1$_{\text{mac}}$$\uparrow$ & P$_{\text{ex}}$$\uparrow$ & R$_{\text{ex}}$$\uparrow$ & F1$_{\text{ex}}$$\uparrow$ \\ \hline
        R2Gen \cite{chen2020generating} &  0.415 & 0.249 & 0.319 & 0.259 & 0.135 & 0.151 & 0.370 & 0.237 & 0.254 \\ 
        R2GenCMN \cite{chen-etal-2021-cross-modal} & 0.423 & 0.256 & 0.341 & 0.257 & 0.134 & 0.152 & 0.373 & 0.291 & 0.309 \\ 
        CvT2DistilGPT2 \cite{nicolson_improving_2023} & 0.275 & 0.255 & 0.337 & 0.261 & 0.135 & 0.153 & 0.376 & 0.294 & 0.313 \\ 
        RGRG \cite{tanida2023interactive} & 0.430 & 0.272 & 0.361 & 0.270 & 0.142 & 0.160 & 0.401 & 0.436 & 0.427 \\ 
        $\mathcal{M}^2$ Transformer \cite{cornia2020m2} & 0.467 & 0.429 & 0.459 & 0.283 & 0.171 & 0.197 & 0.436 & 0.461 & 0.440 \\ \hline
        Ours &  \textbf{0.495} & \textbf{0.515} & \textbf{0.505} & \textbf{0.311} & \textbf{0.256} & \textbf{0.256} & \textbf{0.515} & \textbf{0.503} & \textbf{0.497} \\ \hline
\end{tabular}}
\label{tab:ce}
\end{table}

\begin{table}[!t]
\centering
\setlength{\tabcolsep}{7pt}
\renewcommand{\arraystretch}{0.9}
\caption{Performance comparison between our method and other SOTA methods for attention generation.}
\resizebox{\linewidth}{!}{
\begin{tabular}{l|ccccccc}
\hline
        Methods & fgIoU$\uparrow$ & bgIoU$\uparrow$ & fwIoU$\uparrow$ & SSIM$\uparrow$ & PSNR$\uparrow$ & L1$\downarrow$ & L2$\downarrow$  \\ \hline
        R2Gen \cite{chen2020generating} &  15.87 & 56.03 & 45.08 & 0.35 & 9.12 & 0.840 & 0.200 \\ 
        R2GenCMN \cite{chen-etal-2021-cross-modal} & 18.84 & 64.55 & 51.72 & 0.37 & 10.81 & 0.179 & 0.049 \\ 
        CvT2DistilGPT2 \cite{nicolson_improving_2023} & 17.73 & 66.48 & 48.61 & 0.41 & 11.09 & 0.271 & 0.065 \\ 
        RGRG \cite{tanida2023interactive} & 21.53 & 66.98 & 55.65 & 0.41 & 12.44 & 0.210 & 0.055 \\ 
        $\mathcal{M}^2$ Transformer \cite{cornia2020m2} & 23.19 & 69.06 & 60.22 & 0.45 & 14.51 & 0.120 & 0.031 \\ \hline
        Ours &  \textbf{30.15} & \textbf{89.08} & \textbf{80.69} & \textbf{0.60} & \textbf{17.41} & \textbf{0.084} & \textbf{0.022} \\ \hline
\end{tabular}}
\label{tab:heatmap}
\end{table}

\begin{figure}[!t]
    \centering
    \includegraphics[width=\linewidth]{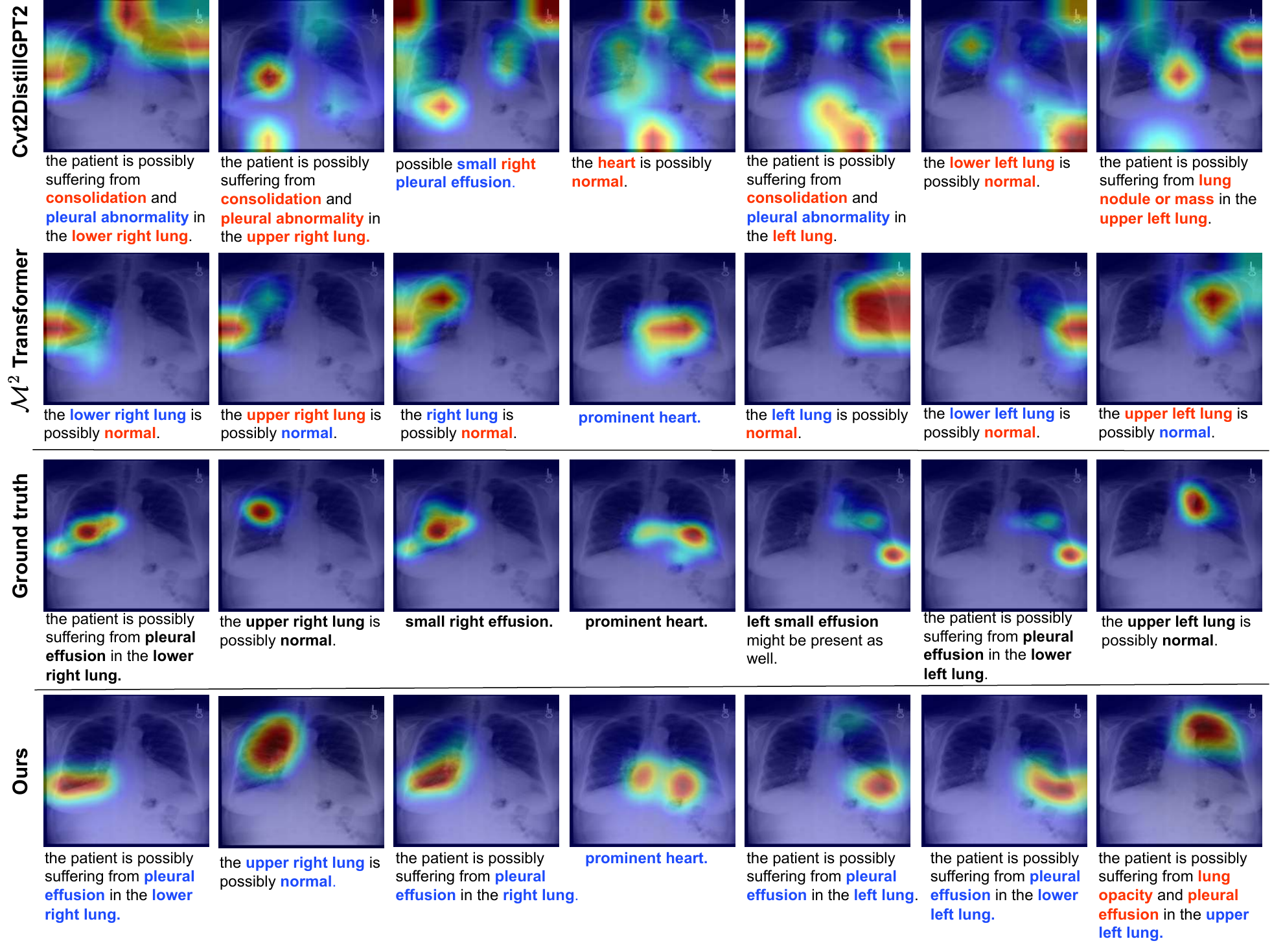}
    \caption{Qualitative comparison. The \textcolor{blue}{blue text} highlights consistent content with ground truth. The \textcolor{red}{red text} indicates incorrect content. }
    \label{fig:qualitative}
\end{figure}

\subsection{Implementation details}

\noindent
\textbf{Architecture details.} GAP comprises an FCN layer for an input patch size of $16 \times 16$ as the Linear Projection, 4 fusion layers, each with the Add Fusion block \cite{pham2024ai}, and the Self-attention block with a hidden dimension of 240 and 6 attention heads. A BiomedCLIP \cite{zhang2023biomedCLIP} is used as CLIP and an MLP with 3 hidden layers of 256 neurons each as the Predictor. The Report Decoder is a GPT2~\cite{radford2019language} initialized with DistillGPT2~\cite{sanh2019distilbert} that has 12 heads, 6 layers, and a hidden dimension of 768. The SAP's Visual Encoder is a CvT~\cite{wu2021cvt} initialized with ImageNet~\cite{deng2009imagenet} and $|\mathcal{V}| \in R^{768}$. We train \model with a learning rate of $5e-5$, batch size of $32$, $6,000$ iterations, and AdamW optimizer \cite{loshchilov2018decoupled}.




\noindent
\textbf{Metrics.} Follow \cite{pham2024ai,tanida2023rgrg,shetty2017speaking}, we evaluate our method based on three criteria: 
\begin{itemize}
    \item Natural Language Generation (NLG) metrics: BLEU (B), METEOR (M), ROUGE-L (R), and CIDEr (C) for matching generated report with the reference report; Div@2~\cite{shetty2017speaking} and R@4~\cite{xiong2018move} for diversity generated reports because we observe that report generation model can suffer heavily from generating only one sentence for all samples.
    \item Clinical Efficacy (CE) metrics: we use all micro-, macro-, and example-based Precision, Recall, and F1 score described in \cite{irvin2019chexpert} because NLG metrics alone are ill-suited for measuring clinical correctness \cite{tanida2023rgrg}.
    \item Attention Similarity metrics: we report all foreground IoU (fgIoU), background IoU (bgIoU), frequency-weighted IoU (fwIoU), Structural Similarity (SSIM), Peak signal-to-noise ratio (PSNR), L1, and L2. Note that, these metrics can also indicate interpretability because when a model's predicted areas of focus closely match those of expert radiologists, it suggests that the model's decision-making process is more understandable and interpretable.
\end{itemize}




\noindent
\textbf{Baselines.}
We compare \model with state-of-the-art methods: R2GenCMN \cite{chen-etal-2021-cross-modal}, R2Gen \cite{chen2020generating}, CvT2DistilGPT2 \cite{nicolson_improving_2023}, $\mathcal{M}^2$ Transformer \cite{cornia2020m2}, and RGRG \cite{tanida2023interactive} on \dataset. For each method, we maintain the default hyperparameters as specified by the authors and train all models on our dataset. For previous works, we use the attention scores of the generated sentences as predicted attention. Since RGRG employs a single vector to represent each region, we use its predicted bounding box for evaluation. 

\subsection{Experimental results}

\noindent
\textbf{Quantitative results.}
\cref{tab:nlg,tab:ce,tab:heatmap} demonstrates the effectiveness of our interpretable approach, which outperforms other methods across all criteria. For example, \model has a high advantage in attention prediction and outshines the runner-up \cite{cornia2020m2} by $+20.47$ in fwIoU. The improvement in attention similarity is understandable as our method is explicitly constrained by the heatmap loss while other methods are not designed for Gaze-Interpretable Report Generation. However, \model also outperforms other black box methods, designed specifically for radiology report generation. For example, in NLG, \model significantly surpasses other models by a large margin, i.e. $+0.671$ on CIDEr metric compared to the runner-up \cite{chen-etal-2021-cross-modal}, and $+0.182$ on Div@2 metric compared to the runner-up \cite{cornia2020m2}. On example-based CE metrics, our model also achieves a higher score of $0.497$, $+0.057$ on F1$_{ex}$ respectively compared to the runner-up \cite{cornia2020m2}. One of possible reason is that previous works are not supervised by radiologist's gaze attention, and thus they fail to learn and use incorrect visual information. This will be further confirmed in \cref{fig:qualitative}, where looking at the wrong location causes the model to fail in producing reliable diagnosis, and in \cref{sec:ablation}, where our model also encounters the same issue when trained with a traditional attention mechanism. 


\noindent
\textbf{Qualitative results.} \cref{fig:qualitative} compares our \model with leading methods, showcasing superior performance in generating precise attention heatmaps and diagnosis reports. CvT2DistilGPT2 produces unreliable and mostly incorrect reports due to inaccurate focus, despite occasional recognition of pleural effusion. On the other hand, the $\mathcal{M}^2$ Transformer often misidentifies lungs as normal, although accurately diagnosing the heart. This highlights the effectiveness of our approach and the crucial role of precise anatomical focus in addressing the Gaze-Interpretable Report Generation challenge.

\begin{table}[!t]
\centering
\setlength{\tabcolsep}{3pt}
\renewcommand{\arraystretch}{1.2}
\caption{Ablation study on applying attention ($\mathcal{T}_{Int}$) to pixels vs. features. }
\label{tab:attention_choice}
\resizebox{\textwidth}{!}{%
\begin{tabular}{l|c|ccc|ccc|ccc}
\toprule
\multirow{2}{*}{\textbf{Settings}}  & \multirow{2}{*}{$\mathcal{T}_{Int}$} & \multicolumn{3}{c|}{\textbf{Attention}} & \multicolumn{3}{c|}{\textbf{NLG}} & \multicolumn{3}{c}{\textbf{CE}} \\ \cline{3-11} 
 &   & \textbf{fwIoU}$\uparrow$ & \textbf{PSNR}$\uparrow$ & \textbf{L1}$\downarrow$ & \textbf{B4}$\uparrow$ & \textbf{C}$\uparrow$ & \textbf{Div@2}$\uparrow$ & \textbf{P}$_{\text{ex}}$$\uparrow$ & \textbf{R}$_{\text{ex}}$$\uparrow$ & \textbf{F1}$_{\text{ex}}$$\uparrow$ \\ \midrule
 \multirow{2}{*}{On pixel} & \xmark & 62.07 & 13.50 & 0.122 & 0.428 & 3.004 & 0.678 & 0.400 & 0.416 & 0.410 \\ 
  & \checkmark & 73.35 & 15.97 & 0.094 & 0.464 & 3.300 & 0.734 & 0.434 & 0.453 & 0.447 \\ \hline
 \multirow{2}{*}{On feature} & \xmark& 79.11 & 17.24 & 0.085 & 0.545 & 3.947 & 0.837 & 0.505 & 0.498 & 0.487 \\
  & \checkmark & 80.69 & 17.41 & 0.084 & 0.561 & 4.026 & 0.854 & 0.515 & 0.503 & 0.497 \\ \bottomrule
\end{tabular}%
}
\end{table}

\begin{table}[!t]
\centering
\setlength{\tabcolsep}{3pt}
\renewcommand{\arraystretch}{1.2}
\caption{Ablation study on the penalty effects.}
\label{tab:penalty_effects}
\resizebox{\textwidth}{!}{%
\begin{tabular}{l|ccc|ccc|ccc}
\toprule
\multirow{2}{*}{\textbf{Settings}} & \multicolumn{3}{c|}{\textbf{Attention}} & \multicolumn{3}{c|}{\textbf{NLG}} & \multicolumn{3}{c}{\textbf{CE}} \\ \cline{2-10}
 &  \textbf{fwIoU}$\uparrow$ & \textbf{PSNR}$\uparrow$ & \textbf{L1}$\downarrow$ & \textbf{B4}$\uparrow$ & \textbf{C}$\uparrow$ & \textbf{Div@2}$\uparrow$ & \textbf{P}$_{\text{ex}}$$\uparrow$ & \textbf{R}$_{\text{ex}}$$\uparrow$ & \textbf{F1}$_{\text{ex}}$$\uparrow$ \\ \midrule
 w/o. penalty & 78.12 & 16.58 & 0.088 & 0.515 & 3.763 & 0.849 & 0.490 & 0.479 & 0.478 \\
 + $\lambda_c$ & 77.41 & 16.74 & 0.088 & 0.534 & 3.871 & 0.837 & 0.500 & 0.493 & 0.483 \\
 + $\lambda_h$ & 79.74 & 17.07 & 0.085 & 0.524 & 3.834 & 0.837 & 0.500 & 0.488 & 0.478 \\
 + $\lambda_c + \lambda_h$ & 80.69 & 17.41 & 0.084 & 0.561 & 4.026 & 0.854 & 0.515 & 0.503 & 0.497 \\ \bottomrule
\end{tabular}%
}
\end{table}

\begin{table}[!t]
\centering
\setlength{\tabcolsep}{3pt}
\renewcommand{\arraystretch}{1.2}
\caption{Ablation study on training with the proposed anatomical gaze attention vs. anatomical segmentation vs. traditional attention. }
\label{tab:3way_training}
\resizebox{\textwidth}{!}{%
\begin{tabular}{l|ccc|ccc|ccc}
\toprule
\multirow{2}{*}{\textbf{Settings}}  & \multicolumn{3}{c|}{\textbf{Attention}} & \multicolumn{3}{c|}{\textbf{NLG}} & \multicolumn{3}{c}{\textbf{CE}} \\ \cline{2-10} 
 &    \textbf{fwIoU}$\uparrow$ & \textbf{PSNR}$\uparrow$ & \textbf{L1}$\downarrow$ & \textbf{B4}$\uparrow$ & \textbf{C}$\uparrow$ & \textbf{Div@2}$\uparrow$ & \textbf{P}$_{\text{ex}}$$\uparrow$ & \textbf{R}$_{\text{ex}}$$\uparrow$ & \textbf{F1}$_{\text{ex}}$$\uparrow$ \\ \midrule
 Traditional attention  & 69.71 & 14.59 & 0.148 & 0.395 & 3.113 & 0.656 & 0.399 & 0.411 & 0.406 \\ 
 Anatomical segmentation  & 79.95 & 17.11 & 0.090 & 0.551 & 3.998 & 0.849 & 0.485 & 0.490 & 0.488 \\ 
 
  Anatomical gaze attention & 80.69 & 17.41 & 0.084 & 0.561 & 4.026 & 0.854 & 0.515 & 0.503 & 0.497 \\ \bottomrule
\end{tabular}%
}
\end{table}

\subsection{Ablation Study}
\label{sec:ablation}
\noindent
\textbf{Applying gaze attention on pixel vs. features.} In \cref{sec:arch}, we apply the predicted gaze attention on the features based on the intuition that the encoder may extract important context information to represent a patch feature besides the patch pixels. For example, the shape or spatial information can be in the feature. This ablation alters only the attention choice, keeping penalty terms and other components as initially proposed. Therefore, we design an ablation study: Instead of applying the gaze attention on the feature, we apply it to the image input, then feed the masked image into the encoder again. Every other setting is kept the same.
The results are shown in  \cref{tab:attention_choice}. Indeed, the findings indicate that incorporating gaze attention directly into the input detracts from model performance. As mentioned in \cref{sec:arch}, this approach can obscure spatial details, leading to confusion. For instance, a heatmap centered on the left lung may not clarify enough whether it targets the left or right side due to uniform coloration. This issue is amplified when training the model without an intention token (w/o. IT), as shown in our table's first row. On the other hand, attending to the latent feature improves the performance, and applying the intention token can slightly boost the performance. 

\noindent
\textbf{Anatomical gaze attention vs. anatomical segmentation vs. traditional attention.} Interpretable model is often mistakenly thought to be harmful to the performance~\cite{rudin2019stop}. To demonstrate that this is not the case for our proposed model. We design an ablation study with two more settings:
\begin{itemize}
    \item \textit{Traditional attention}. We remove the Gaze Attention Predictor (GAP) module and instead use a simple self-attention module to flexibly weigh the importance of every patch feature. In other words, we let the model automatically decide the importance of every patch. 
    \item \textit{Anatomical segmentation}. Instead of supervising our model on gaze attention ground truth, we supervise the GAP with our anatomical segmentation masks. Then we mask out patches that are not in the predicted mask. In other words, we let the GAP module be an anatomical mask predictor, and a patch is important, i.e. its weight is 1.0 if it is inside the anatomy of interest.
\end{itemize}
\cref{tab:3way_training} shows that our proposed anatomical gaze attention supervision is effective and outperforms other settings. For the traditional attention setting, the black-box and unconditional training pipeline causes the model to not know where to look, i.e. low scores on Attention criteria, and hence, it fails to give satisfaction diagnosis, i.e. low NLG and CE scores. One of the possible reasons for this is because self self-attention mechanism is well-known for its data-hungry nature~\cite{khan2022transformers}. On the other hand, training the GAP module with segmentation masks slightly decreases the performance. One possible reason is that we let the model use too much information, which can confuse the model in some cases. The gain from correctly weighing important patches further confirms our hypothesis in \cref{sec:intro}. Moreover, this suggests that our framework can also be used for segmentation prediction. 

\noindent
\textbf{Effectiveness of penalty terms.} The intuition behind the penalty terms is simple, yet effective. We design an ablation study: we train the model without penalty terms to demonstrate the effect of every penalty. As a result, we find that our penalties based on the idea of looking at the correct anatomy are beneficial to the model, as shown in \cref{tab:penalty_effects}.


\section{Conclusion} 

In this work, we have introduced \dataset, a curated dataset for gaze interpretable radiology report generation. Our dataset contains CXR images with aligned gaze sequence, gaze attention heatmap, and the reports associated with seven anatomical parts of the lung. We then presented a novel method for generating descriptive reports of chest X-ray images, using heatmaps based on radiologist annotations to focus the model's attention and reduce the likelihood of misinterpreting irrelevant regions. Our main contribution is the successful application of a radiologist-informed attention mechanism that guides a generative model, thereby enhancing the accuracy, reliability, and interpretability of automated CXR report generation. We hope that the release of our dataset will advance more research on interpretable report generation. 

\begin{credits}
\subsubsection{\ackname} This material is based upon work supported by the National Science Foundation (NSF) under Award No OIA-1946391, NSF 2223793 EFRI BRAID, National Institutes of Health (NIH) 1R01CA277739-01.
\end{credits}



%
%
%

\clearpage
\bibliographystyle{splncs04}
\bibliography{refs}
\end{document}